\def\year{2019}\relax
\documentclass[letterpaper]{article} 
\usepackage{aaai19}  
\usepackage{times}  
\usepackage{helvet}  
\usepackage{courier}  
\usepackage{url}  
\usepackage{graphicx}  
\title{The Concept of Criticality in Reinforcement Learning}
\author{%
Yitzhak Spielberg \and Amos Azaria\\
Ariel University, Israel
}

\usepackage[section]{placeins} 
\usepackage{subcaption}
\usepackage{tikz}
\usepackage{amsmath} 
\usepackage{amsthm}
\usepackage{algorithm,algorithmic}
\usepackage{changepage}   
\tikzset{
	treenode/.style = {shape=rectangle, rounded corners,
		draw, align=center,
		top color=white, bottom color=blue!20},
	root/.style     = {treenode, font=\Large, bottom color=red!30},
	env/.style      = {treenode, font=\ttfamily\normalsize},
	dummy/.style    = {circle,draw}
}
\begin{document}
\maketitle
\begin{abstract}
Reinforcement learning methods carry a well known bias-variance trade-off in n-step algorithms for optimal control. Unfortunately, this has rarely been addressed in current research. 
This trade-off principle holds independent of the choice of the algorithm, such as n-step SARSA, n-step Expected SARSA or n-step Tree backup. A small n results in a large bias, while a large  n leads to large variance. 
The literature offers no straightforward recipe for the best choice of this value. While currently all n-step algorithms use a fixed value of n over the state space we extend the framework of n-step updates by allowing each state to have its specific n.
	
	 We propose a solution to this problem within the context of human aided reinforcement learning. Our approach is based on the observation that a human can learn more efficiently if she receives input regarding the criticality of a given state and thus the amount of attention she needs to invest into the learning in that state. 
This observation is related to the idea that each state of the MDP has a certain measure of criticality which indicates how much the choice of the action in that state influences the return. In our algorithm the RL agent utilizes the criticality measure, a function provided by a human trainer, in order to locally choose the best stepnumber n for the update of the Q function.
\end{abstract}
\section{Introduction}
Reinforcement learning based methods have recently shown great success in many domains, including Atari games \cite{Mnih}, Go \cite{Silver}, and autonomous vehicles \cite{Shalev,Sallab}. However, the bias-variance trade-off in n-step TD algorithms,such as n-step SARSA, n-step Expected SARSA and n-step Tree Backup, hasn't been addressed that much in recent research. 
This effect is inherent to  the stochastic nature of Markov Decision Processes (MDP) and invariant to the specific reinforcement learning  algorithm. The bias-variance tradeoff rule states that a small n leads to large bias, whereas big n results in a high variance of the n-step bootstrap update of the Q function \cite{SutBar,DeAsis}. The variance phenomenon is an inherent property of the n-step update which contains a sum of n individual rewards. It is a consequence of the fact that the variance of a sum of random variables grows with the amount of summands.
Conversely, the bias, which comes from the possibly biased current estimate of the Q-function in the target state, is a decreasing function of n; This can be easily seen by considering the extreme case of the maximal possible n which is equivalent to Monte Carlo updates; these updates are completely unbiased since they don't involve any estimates of the Q-function. 

Currently there are very few instructions for choosing the best value of $n$ in reinforcement learning literature. 
We address this challenge in inspiration by the way humans teach each other. 
One of the ways in which a person can assist another in learning, is by indicating which situations are more critical and as such require higher attention. If, for example, a student driver approaches an obstacle in the road, her teacher may state to her that she must watch out, without suggesting exactly which action to take (e.g. slowing down, turning the wheel right or left etc.). 
If the car later hits that obstacle, the student will understand that it probably took a wrong action back when the teacher has warned her. 

\par This observation translated into the formal language of reinforcement learning might motivate the introduction of the concept of criticality. We  can think of the criticality of a state as a measure which indicates how much the choice of the action in that state influences the return. The criticality of a given state $h(s) \in [0,1]$ could be provided by either by a rule of thumb, or explicitly by a human trainer. Technically speaking we can think of the criticality of a state as being proportional to the variance of the optimal Q function $Q^*$ wrt. the possible action choices in that state.

\section {Related Work}
Multi-step TD methods, such as the n-step SARSA and SARSA($\lambda$) create a spectrum of algorithms where at one end exist one-step TD learning, and at the other - Monte Carlo Methods. All of these algorithms use the n-step return which is subject to bias-variance tradeoff. Various approaches have been developed to tackle this challenge.
  
De Asis \cite{DeAsisPer} addresses this problem for off-policy n-step TD methods, such as n-step Expected SARSA, via the introduction of so called \textit{control variates}. These special terms have the impact of an expectation correction. Therefore they can be used to decrease the bias of the n-step return.

Jiang et. al. \cite{JiangDoub} propose an alternative solution for this problem for the prediction task (not the optimal control task). They introduce an unbiased estimator which corrects the current estimate  of the value function $\hat{V}(S_t)$. This estimator is robust in the sense that it remains unbiased even when the function class for the value function is inappropriate.

Richard Sutton et.al. \cite{SutEmp} suggest an improvement of TD($\lambda$)  which achieves an effective bias reduction for the updates. This beneficial effect is a consequence of specific weights  which are being assigned to any given update of the value function. The proposed variant of TD($\lambda$) is particularly useful for off-policy learning, where ordinary TD($\lambda$) suffers from a deficit of stability.

Unlike all of the above mentioned approaches our method does not manipulate the updates of the (action-)value function a-posteriori; instead of this, it chooses the appropriate stepnumber for the update a-priori. This is done by using the criticality function, which is closely related to the  variance of the update. Therefore, in our case, we can speak about a technique, which speeds up the learning by controlling the variance of the updates.

\section {The  relation between criticality and the stepnumber}
All prominent n-step RL algorithms, such as n-step SARSA, n-step Expected SARSA and n-step Tree Backup, use a fixed stepnumber $n$ for bootstrapping which stays constant both in in the course of an episode and during the complete learning process. 
In our approach we use a varying stepnumber which is specific to each state encountered during an episode. We believe that the concept of the criticality of a state allows the determination of the optimal $n$ for a given state.

The intuition behind this idea is very straightforward. To develop it we present a simple example. 
Let us assume that in our environment most of the states have only one available action, and that there is no randomness in the MDP, that is, a given state action pair determines the next state.
Let us assume, that during the learning process the agent encounters some sequence of states $\{S_0,S_1,S_2,S_3\}$ of which only $S_3$ has multiple actions available. In this situation, obviously $S_0,S_1,S_2$ should be assigned a criticality of $0$ (since the agent has no choice, and therefore its ``choice'' has no influence on the final reward) whereas for simplicity we will assign to $S_3$ a criticality of $1$. Clearly, whenever the agent arrives at $S_0$, the next states it visits will always be ($S_1,S_2,S_3$). We would like to determine which of the states ($S_1,S_2$ or $S_3$) should be used as the update target for $S_0$. Consider the simple 1-step SARSA. This algorithm will update $S_0$ towards $S_1$ and in the next step $S_1$ towards $S_2$. These updates will be repeated in each episode where these states are being visited so it is easy to see that asymptotically $S_0$ will be updated towards $S_2$. Therefore there is no benefit from selecting $S_1$ as the update target for $S_0$ versus selecting $S_2$, and the selection of $S_2$ may speed up the convergence.  
Using the same argument we can conclude that $S_3$ is a better update target than $S_2$.

\section{CVS}
The presented example may lead to the conclusion that the update target for a given state should be the next future state which has a criticality of $1$. However, how will this idea work out if none of the states has a criticality of $1$? Our actual approach, which is more robust will use cumulative criticality. That is: we will sum up criticality over the encountered states and postpone our update until the criticality accumulates to one. We easily see that in our above example this will result in exactly the same update target as the simple strategy that doesn't use accumulation. This method will produce large stepnumbers in uncritical domains of the state space and therefore we can expect a speed-up in learning. We will call this algorithm "Criticality-based 
Varying Stepsize" (CVS).
\begin{algorithm}[H]
	\caption*{The CVS algorithm }
	\begin{algorithmic}
		\STATE $CrtCum(s,a)=0$ for all s,a (cumulative criticality) \\
		WaitList= $\lbrace \rbrace$ (states waiting for update) \\
		pick initial state $S=S_0$ and action $A=A_0$ greedily \\
		while $S \neq Terminal$ \\
		\begin{adjustwidth}{.5cm}{}
			add $(S,A)$ to WaitList \\
			observe $R,S'$ \\
			pick $A'$ greedily \\
			for $(\hat{S},\hat{A})$ in WaitList \\
			\begin{adjustwidth}{.5cm}{}
				
				if $CrtCum(\hat{S},\hat{A}) \geq 1$: \\
				\begin{adjustwidth}{.5cm}{}
				 update $Q(\hat{S},\hat{A})$ with update target $(S',A')$ \\
				 delete $(\hat{S},\hat{A})$ from WaitList \\
				 $CrtCum(\hat{S},\hat{A})=0$ \\
				\end{adjustwidth}{}
				else: \\
				\qquad $CrtCum(\hat{S},\hat{A})+=h(S')$ \\
			\end{adjustwidth}{}
			S,A = S',A'		
			
		\end{adjustwidth}{}
		for $(\hat{S},\hat{A})$ in WaitList \\
		\qquad update $Q(\hat{S},\hat{A})$ \\
		\qquad with update target $(S',A')$
		
	\end{algorithmic}
\end{algorithm}

\section{Evaluation of CVS in the Road-Tree environment}
In this section we introduce the Road-Tree environment, an environment which is particularly appropriate to understand the benefits of CVS. 
We test the algorithm against a number of widely used reinforcement algorithms in order to prove it's efficiency. By default, if not specified otherwise, we won't discount the reward ($\gamma=1$)
and our initial Q-function will be constant over the state-action space. Our default values are 
$\epsilon=0.1$, and
$\alpha=0.1$.
\subsection{The Road-Tree environment}
\begin{figure}
	\centering
	\begin{tikzpicture}
	[
	level distance          = 5em,
	level 1/.style={sibling distance=6em},
	level 2/.style={sibling distance=2em},
	edge from parent/.style = {draw, -latex},
	every node/.style       = {font=\footnotesize},
	sloped
	]
\node [root] {0}
	child { node [dummy] {0}      
		child { node [dummy] {0}
			edge from parent node  {10} }
		child { node [dummy] {7}
			edge from parent node  {15} }
		edge from parent node  {20} }
	child { node [dummy] {1}
		child { node [dummy] {1}
			edge from parent node  {15} }
		child { node [dummy] {1}
			edge from parent node 
			{15}}
		edge from parent node  {10} };
	\end{tikzpicture}
	\caption {Road-Tree example. The number in a critical state (junction or terminal state) represents the reward in that state. The number on an edge is the distance between the corresponding nodes.}
	\label{tree0}
\end{figure}
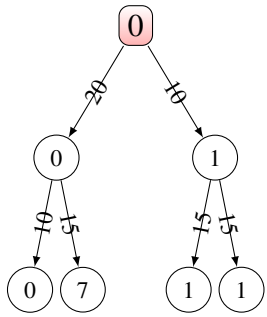

In order to test CVS, we construct a plain environment, named Road-Tree, which has a natural criticality function corresponding to it. Road-Tree, is tree-like structured. The agent starts at the root and always moves in one direction--downward. There are two types of states. In a simple state there is only one possible action. In a junction state the agent needs to choose between multiple roads. The reward upon stepping onto a simple state is always zero. The reward is nonzero only upon reaching a junction or a terminal state. Moreover the reward may vary across junctions and terminal states. Figure \ref{tree0} illustrates a simple Road-Tree environment. The numbers in the junctions represent the rewards. The numbers on the edges show the distance between the two corresponding junctions, that is the number of simple states between them (a distance of $n$ indicates $n-1$ simple states).

The very natural criticality function which we are going to use in the Road-Tree environment assigns zero to a simple state and one to a junction or terminal state.\\

\subsection{CVS vs. Q-Learning}
We now compare the performance of CVS against that of Q-Learning in the 2-level Road-Tree from fig. \ref{tree0}. Clearly, the optimal policy is defined by initially going to the left and then to the right, ending up at the terminal state that has a reward of $r=7$. 
In Q-Learning, due to the relatively big distance between the intermediate junction that has a reward of $r=0$ and the optimal terminal state, the optimal reward ($r=7$) will be backpropagated to the intermediate junction very slowly. The other intermediate junction that has a reward of $r=1$ will be much more attractive to the agent and therefore, the agent, might remain in that nonoptimal path for a long period of time. Conversely, the CVS agent will backpropagate the optimal reward terminal state to the intermediate junction immediately after the fist visit and therefore should quickly converge to the optimal policy. 
\par
\begin{figure}
	\centering
	\includegraphics[scale=0.5]{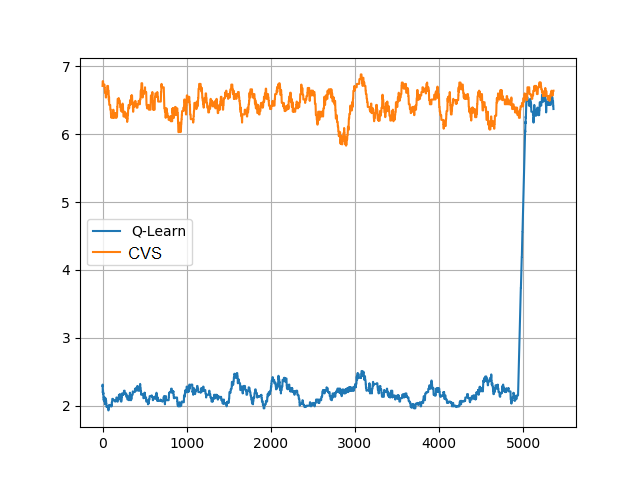}\\
	\caption {CVS against Q-Learning in a 2-level  Road-Tree environment (Figure \ref{tree0}). The x-axis is the episode number; the y-axis corresponds to the average return over the last 100 episodes. CVS converges to the optimal policy immediately, while Q-Learning requires about 7000 episodes to converge. }
	\label{roadtree_qlearn}
\end{figure}

 The plot on fig. \ref{roadtree_qlearn} confirms our elaboration. The Q-Learning agent needs about 7000 episodes to converge to the optimal policy; the CVS agent, by contrast, converges almost immediately. 
\subsection{CVS vs. Watkin's Q($\lambda$)}
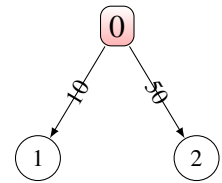
\begin{figure}
	\centering
	\begin{tikzpicture}
	[
	level distance          = 5em,
	level 1/.style={sibling distance=6em},
	edge from parent/.style = {draw, -latex},
	every node/.style       = {font=\footnotesize},
	sloped
	]
	\node [root] {0}
	child { node [dummy] {1}      
		edge from parent node  {10} }
	child { node [dummy] {2}
		edge from parent node  {50} };
	\end{tikzpicture}
	\caption {Plain Road-Tree Environment}
	\label{tree1}
\end{figure}

We now compare the performance of CVS against that of Watkin's Q($\lambda$) in a very simple Road-Tree environment (see Figure \ref{tree1}). Watkin's Q($\lambda$) is a very popular state of the art algorithm in reinforcement learning, which makes use of eligibility traces. 
\par In our simple Road-Tree example (fig. \ref{tree1}) the right branch has a higher terminal reward and therefore is clearly the optimal policy. However we can easily see that Q($\lambda$) for this case, $\lambda<1$, might have a hard time finding this optimal policy. For a given $\lambda<1$ the amount of episodes required to backpropagate the terminal reward to the root is an increasing function of the road's length. Therefore the higher reward of the optimal branch will be backpropagated towards the root much slower than the lower reward of the non-optimal branch. \\
This leads to 2 possible scenarios in Q$(\lambda)$ learning. The first one, which is  very unprobable, occurs when before visiting the nonoptimal branch for the first time the optimal branch has been visited so often that the terminal reward has been backpropagated to the root to a degree, sufficient to outperform a one-time visit of the non-optimal branch. The second one, that is much more probable, is that Q($\lambda$) will be stuck in the non-optimal branch once it has traveled it . Certainly due to epsilon-greedy exploration eventually $Q(\lambda)$ will converge to the optimal policy, however for a small $\epsilon$ this might take a very long time. 

\par
\begin{figure}
	\centering
	\includegraphics[scale=0.9]{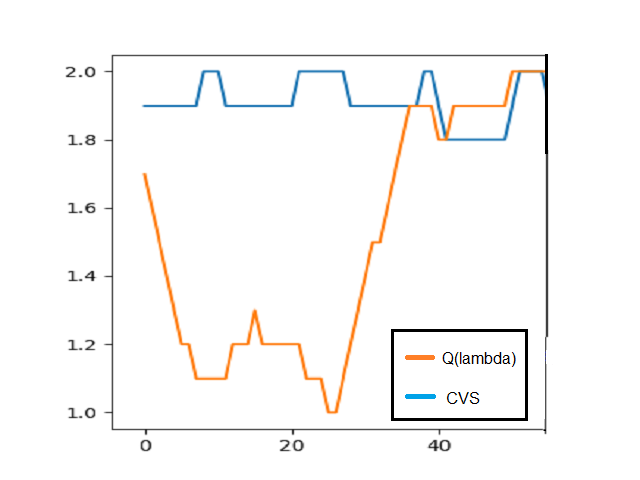}\\
	\caption {A comparison of CVS against Q($\lambda$) in the simple Road-Tree environment (Figure \ref{tree1}). The x-axis is the episode number; the y-axis corresponds to the average return over the last 10 episodes. CVS converges to the optimal policy immediately while Q($\lambda$) requires about 40 episodes. }
	\label{plot_1}
\end{figure}

 In order to validate our elaboration we applied both learning algorithms Q($\lambda$) and CVS to learn the optimal policy for our Road-Tree environment.  The initial Q-function and the value of $\lambda$  were set to: 
$Q_0=0$ 
$\lambda=0.9$. 
For each episode we recorded the return. In order to smoothen the plot, we applied a running average with a width of 10 episodes to the return. The plot ( fig. \ref{plot_1}) shows the smoothened return vs the episode number. Indeed the plot proves our initial assumption. CVS instantly chooses the optimal  policy. The variations in the return are caused only by the ongoing exploration.  $Q(\lambda)$, however,takes about 40 episodes until the policy becomes optimal.
\par
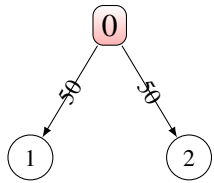
\begin{figure}
	\centering
	\begin{tikzpicture}
	[
	level distance          = 5em,
	level 1/.style={sibling distance=6em},
	edge from parent/.style = {draw, -latex},
	every node/.style       = {font=\footnotesize},
	sloped
	]
	\node [root] {0}
	child { node [dummy] {1}      
		edge from parent node  {50} }
	child { node [dummy] {2}
		edge from parent node  {50} };
	\end{tikzpicture}
	\caption {simple Road-Tree Environment}
	\label{tree3}
\end{figure}

 We now compare the performance of CVS against that of Q($\lambda$) with a slightly more challenging environment, in which the distances between the root and the terminal states are equal(fig. \ref{tree3}). 

\begin{figure}
	\centering
	\includegraphics[scale=0.8]{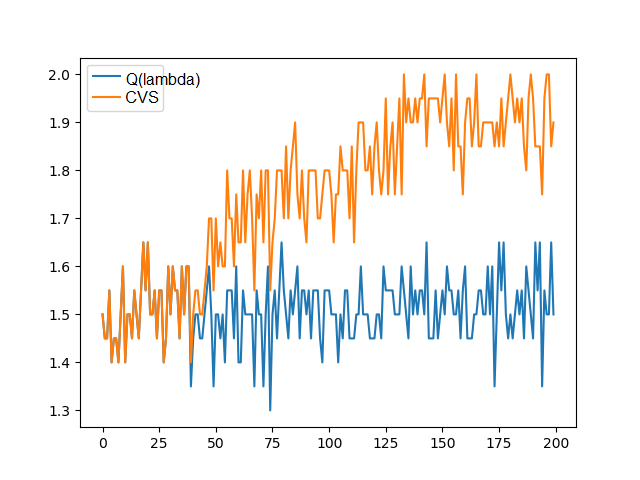}\\
	\caption {CVS performance against Q($\lambda$) in a more challenging Road-Tree environment. The x-axis is the episode number; the y-axis corresponds to the average return over the last 20 episodes. Q($\lambda$) struggles to make any progress; in contrast, CVS takes about 200 episodes to converge to the optimal policy.}
	\label{plot_2}
\end{figure}

Intuitively, as soon as the CVS agent will see the optimal policy, the corresponding Q-value will become equal to $q*$ which basically terminates the learning process. We validated our intuition experimentally via 20 simulations, each containing a 200 episodes long training session and averaged the returns over the simulations.  As might have been expected the plot on fig. \ref{plot_2} confirms our guess. We see, that in fact CVS converges to the optimal policy much quicker than Q($\lambda$).
\subsection {CVS vs Monte Carlo}
\begin{figure}
	\centering
	\begin{tikzpicture}
	[
	level distance          = 5em,
	level 1/.style={sibling distance=6em},
	level 2/.style={sibling distance=2em},
	edge from parent/.style = {draw, -latex},
	every node/.style       = {font=\footnotesize},
	sloped
	]
	\node [root] {0}
	child { node [dummy] {0}      
		child { node [dummy] {0}
			edge from parent node  {} }
		child { node [dummy] {1}
			edge from parent node  {} }
		edge from parent node  {} }
	child { node [dummy] {1}
		child { node [dummy] {-2}
			edge from parent node  {} }
		child { node [env] {...}
			edge from parent node  {} }
		child { node [dummy] {-2}
			edge from parent node  {} }
		child { node [dummy] {1}
			edge from parent node 
			{10}
		}
		edge from parent node  {10} };
	\end{tikzpicture}
	\caption {Road-Tree Environment with two levels and a large number of branches. The optimal branch is "hidden" among a multitude of suboptimal branches that share the same parent. }
	\label{tree2}
\end{figure}
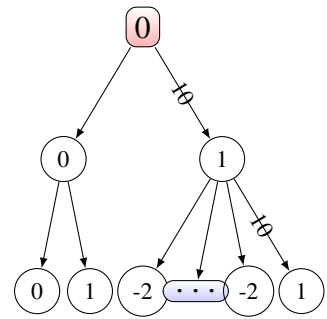

In the previous section we presented an example in which CVS outperformed Q($\lambda$). An interesting observation is that in that particular example CVS functioned exactly like Monte Carlo (MC); the update targets were always the terminal states. This observation immediately raises the question whether CVS is able to outperform MC. We are going to show that there is in fact a situation,  where it is the case. \\
In the current example (fig. \ref{tree2}) we have a slightly more complex Road-Tree
than in the previous section: a 3-level tree. On the second level there are two junctions with different rewards.The junction with the higher reward, $s_{high}$, has many  children. 
All of these children with the exception of one child have negative rewards. The only child with the positive reward corresponds to the optimal policy. The second junction on level 2, which we denote with $s_{low}$, has a lower reward. It has two children with nonnegative rewards. \\
Let us consider how a Monte Carlo agent will act in this environment. The first time  the agent visits $s_{high}$ it is most likely continue to a low reward child, simply because most of the branches have a low reward. Therefore, the total return of the episode will be negative. Once this negative return has been backpropagated to the root, which happens instantly in the case of MC, the agent will avoid to explore policies which pass through $s_{high}$. Some exploration will still take place, but it will be only due to the $\epsilon$.
Therefore, it might take a long time before the agent visits the optimal trajectory which passes through the $s_{high}$ junction. It might take even much longer until it visits this trajectory sufficiently many times until the Q value at the root, for the right action  (towards $s_{high}$) will become higher than that of the left action (towards $s_{low}$).

CVS should learn faster in this Road-Tree. Consider the trajectories which contain $s_{high}$. The Q-value for $s_{high}$ at the root will have $s_{high}$ as the update target. Therefore, it will most of the time choose this branch. Now let us assume the worst case, in which it takes a large amount of episodes until the agent sees the high-reward child of $s_{high}$ for the first time. Certainly this might lead to a situation where $Q(s_{high},a)<Q(s_{low},b)$ for all possible actions $a,b$. However because the learning rate $\alpha$ is small the  difference $\max_b Q(s_{low},b)- \max_a Q(s_{high},a)$ will grow slowly. Since the reward upon reaching $s_{high}$ is higher than that of $s_low$, as long as the aforementioned difference is not too large, the agent will prefer $s_{high}$. Therefore, it should take CVS less episodes to learn the optimal trajectory.

\begin{figure}
	\centering
	\includegraphics[scale=0.8]{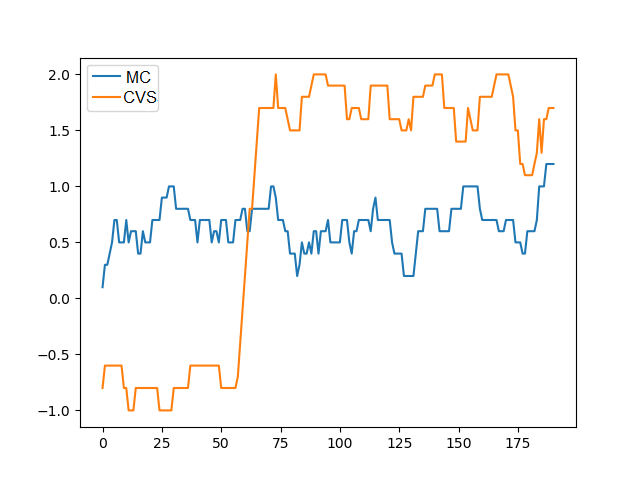}\\
	\caption {MC vs. CVS in a specific Road-Tree environment. Average return over last 10 episodes is plotted against the episode number. MC struggles to make any progress at all, whereas CVS quickly converges to the optimal policy.}
	\label{mc_vs_avs}
\end{figure}

A comparison between MC and CVS is shown in fig. \ref{mc_vs_avs}. From the plot we can imply that, as expected, MC visits $s_low$ most of the time and as a consequence fails to identify the optimal policy. In contrast, CVS visits $s_high$ much more frequently. The plot shows that the optimal policy is executed for the first time after about 60 episodes and from there on CVS mostly keeps following it.

\section{CVS vs. Q-Learning in the Shooter environment}
In this section we describe the performance of CVS versus Q-Learning in a different environment: the Shooter environment. Just like the Road-Tree environment, the Shooter environment can be naturally associated with a simple criticality measure. 

\subsection{The Shooter environment}
The Shooter environment is located on a rectangular playing field of 10x20 (width x length) parcels. This playing field contains multiple objects: a gun, which is located in the first column and whose random position may change from game to game; a bullet, which initially is located at the gun's position; and a moving target, which is located in the last column. Each of these objects occupies exactly one parcel. Furthermore there exists an obstacle of a size of 3 parcels in the 8th column. 
At the beginning of the game the target has a random position in the last column of the field and a random direction of movement which can be either up or down. In every step the target moves by exactly one parcel inside the last column. The direction of the movement is inherited from the previous step with the exception of the case when it hits the wall; in that case the direction is simply being reflected. The agent controls the gun. At any given state of the game the agent can choose one of four actions: Either not shoot at all or shoot in one of the three possible directions - diagonally up, diagonally down or horizontally. The three shooting actions shoot a bullet only if the agent has a bullet to shoot, otherwise these actions are equivalent to doing nothing. 
At any given step the bullet will move by one parcel in the direction it was shot; when hitting a wall it's vertical direction is being reflected; if it hits the obstacle the game is being terminated with a reward of -1; in the case it reaches the last column, the game is being terminated with a reward of +1, if it hits the target or -1, if it does not hit it. 
\begin{figure}
	\centering
	\includegraphics[scale=0.6]{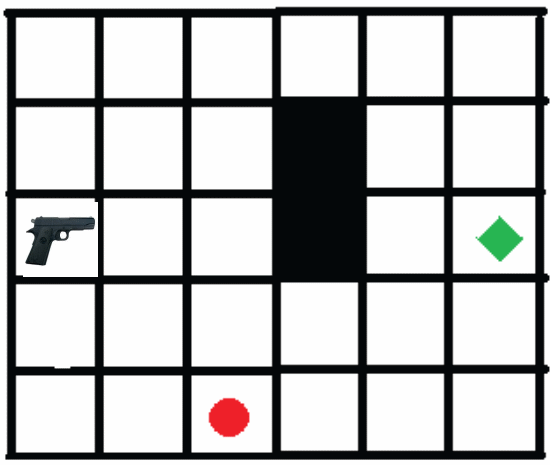}\\
	\caption {Shooter environment. The gun represents the agent's location, the red circle is the bullet, and green diamond represents the target, and the blacked-out squares represent an obstacle. The target is moving, the obstacle isn't. 
    Field size is not the same as in the actual environment. 
    }
	\label{Shooter_env}
\end{figure}
\par There exists a rather natural criticality measure for the Shooter environment. The agent's actions are relevant only before the shot. Moreover before the shot any state can be considered as equally critical. Therefore the most obvious criticality will be binary. It will assign a criticality of 1.0 to any state in which the shot didn't take place yet; and a criticality of 0.0 to any state that occurs after the shot.
\subsection{The performance of CVS vs. Q-Learning}
\begin{figure}
	\centering
	\includegraphics[scale=0.5]{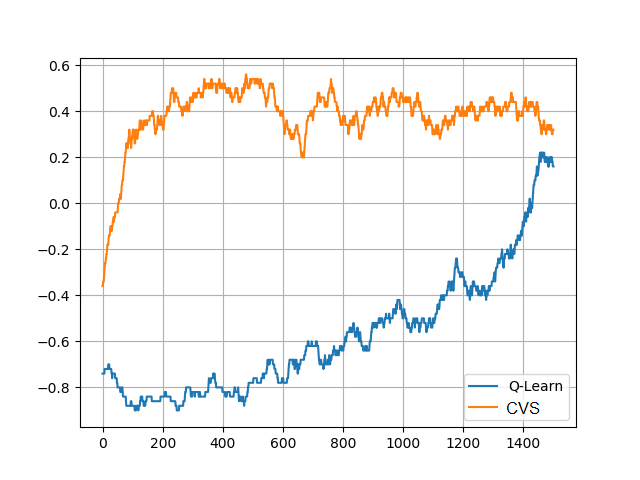}\\
	\caption {A comparison between the performance of CVS against Q-Learning in the Shooter environment. The x-axis shows the episode number; the y-axis corresponds to the average score over the last 100 episodes. }
	\label{plot_avs_shooter}
\end{figure}

In order to compare CVS to Q-Learning, we implemented a tabular Q-Learning agent and a tabular CVS agent. For both agents, we initiated the Q-function to a value of $Q(s)=0$ in every state. The exploration parameter $\epsilon$ was set to a value of 0.1 and remained constant throughout the learning process. The performance of both agents, which was monitored by averaging the scores over the last 100 episodes, is plotted in fig. \ref{plot_avs_shooter}. As depicted in the plot, CVS clearly outperforms Q-Learning. The Q-Learning agent struggles to make any progress during the first 500 episodes; and it takes the Q-Learning agent about 1400 episodes to reach an average score of 0.0. Conversely, the CVS agent reaches an average score of 0.0 already after about 100 episodes, and after 200 episodes it converges to a performance level of 0.4. 

\section {CVS vs DDQN in the Tennis environment}
\subsection{The Tennis environment}
In this section we test CVS performance in the context of Deep-Q-Learning. For this purpose we implemented the Tennis environment which can be associated with a binary criticality function in a very natural way. 
The Tennis environment consists of two rackets (one controlled by the agent and the other by a computer opponent), a ball, and a playing field which has a size of 20x40 (width x length) pixels. 
On this field both the agent's and the opponent's racket occupy one pixel each, in the second and second to last columns. The movements of each racket are defined by the three primitive actions (up, down, stay) which either move the racket by one pixel in the corresponding direction or let it remain at the same position. If the racket is located at the wall, and therefore is not able to move in one of the two directions, executing this action is equivalent to staying at the same position. The ball occupies a single pixel and can move in six directions:[horizontally, diagonally up, diagonally down] either towards the agent or towards the opponent.
If the ball hits either a wall or a racket its direction of movement is reflected. The opponent's policy in the Tennis environment is a noisy variant of the optimal policy. At any given state the opponent chooses the optimal action 
with a probability of $p=0.8$ or some random action with the probability of $q=1-p$. Each game consists of a single point. The agent receives a reward of +1 when it scores, and a reward of -1 when the opponent scores. The starting position of the ball is always at the center of the field. The starting direction is always towards the agent. The exact direction (horizontally, diagonally up, diagonally down) is random.

\subsection{The DDQN algorithm }
We implemented the DDQN algorithm (double DQN) that has been proposed by Hasselt et. al.  \cite{Hass}. The main benefit of this approach over plain DQN is that the second neural net improves the stability of the learning procedure. The strategy to approach the exploration vs. exploitation challenge consists of three learning periods. The first 2000 games are an "exploration-only period". Afterwards we perform a linear decay of the exploration parameter $\epsilon$ which starts at the value $1.0$ and is finally being decreased to the value of $\epsilon_{fin}=0.1$ by the 12000th game. In the final learning period $\epsilon=\epsilon_{fin}$ is constant. Our learning rate is $\alpha=0.0001$ and our reward decay parameter is $\gamma=0.99$. Our neural net takes the 20x40 image as the input and has an output layer whose size equals the amount of possible actions ( in our case three). It has a compact architecture with only three hidden layers: two convolutional and one fully connected layer. The exact structure is [(Conv,32),(Conv,64),(FC,256)]. 
\subsection{The criticality measure}
Since the agents actions are irrelevant when the ball is moving towards the opponent, it is a rather straightforward strategy to set the criticality of these states to zero. When the ball is moving towards the agent there is a variety of options for a meaningful criticality measure; we simply assign criticality of 1.0 to these states. 

\begin{figure}
	\centering
	\includegraphics[scale=0.5]{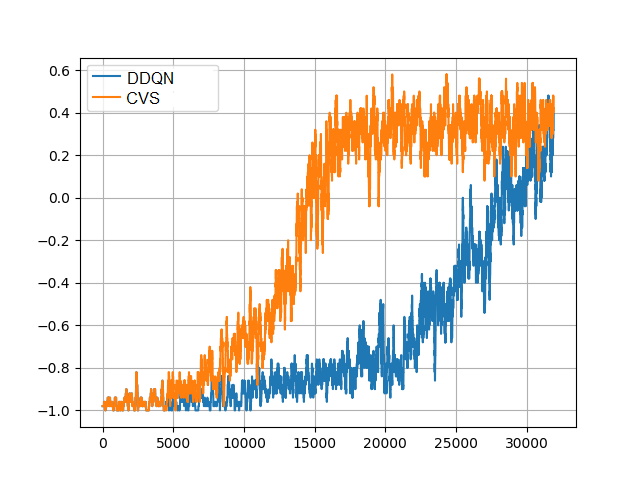}\\
	\caption {DDQN vs CVS in the Tennis environment:  the x-axis corresponds the episode number; the y-axis is the average score over the last 100 episodes. CVS reaches machine level performance (score=0) about two times as fast as DDQN.}
	\label{pong_avs}
\end{figure}

\subsection{Performance of CVS vs DDQN}
The difference between CVS and DDQN is reflected in the update targets. Whereas in DDQN each example in the experience buffer has the next state as the update target in CVS the update target is being chosen according to the CVS algorithm.
We chose to monitor the learning procedure for each algorithm by looking at the average score over the last 100 points. In fig. \ref{pong_avs} the performance boost of CVS in comparison to DDQN is clearly recognizable; after the first 5000 games CVS has only a tiny lead; by game 10000 the lead is already clearly visible; and after 15000 games it becomes significant. The most important observation is that CVS reaches machine level performance about two times faster than DDQN.

\section{Discussion}
All experiments presented in this paper show that CVS outperforms all other baselines which do not take the concept of criticality into account, in terms of convergence speed. However, obtaining a criticality function is not always a trivial task. While a human trainer may provide criticality levels to some states, obtaining a function that evaluates criticality of all states (in a satisfactory matter) may be domain dependent. Furthermore, it may not always be obvious which states should be considered critical and which states should be considered as non-critical. For example, a car driving on a straight road with no traffic may seem as being in a non-critical state. However, a driver that suddenly turns the wheel right (or left), may result in hitting a wall (and resulting with a negative reward), this implies that the state was in fact a critical state. Therefore, in some domains it may be required to refine the concept of criticality in order to improve agent performance in these domains. One such approach may include giving a higher weight to more plausible actions, perhaps by taking the current behavior of the agent into account.

\section{Conclusions and future work}

In recent years most human-aided reinforcement learning approaches improved the agent's performance by either integrating human demonstrations into the learning procedure or augmenting the reward function with human feedback. In this paper we introduce a novel idea which opens an alternative way of human assistance in the agent's learning process: the concept of the criticality of a state.
We implemented the Criticality-based Varying Stepnumber (CVS) agent which uses the concept of criticality in order to locally choose the appropriate stepnumber for the update of the Q-function. 
\par  
We tested the CVS agent in multiple environments, including Road-Tree, the Shooter game and the Tennis game. The conducted experiments showed that CVS is able to outperform  popular reinforcement learning algorithms as Q-Learning, Deep-Q-Networks and Monte Carlo Learning.
\par 
There is a number of promising research directions in the area of criticality based algorithms. The first is the development of interfaces which would enable the human trainer to communicate criticality related information to the agent. Such a step would be a crucial towards a practical realization of Criticality-based learning. An alternative research direction could arise from the fact that our experiments were limited to criticality functions which were provided by the human trainer for each state of the MDP. In more complex environments it might be interesting to consider the option, that the criticality function will be provided by the human only on a certain portion of the states and will be generalized to all other states by supervised learning techniques. Such an approach could significantly reduce the required effort for the human trainer and consequentially make Criticality-based learning much more attractive.
\newpage

\bibliography{mybib}{}

\begin{thebibliography}{}

\bibitem[\protect\citeauthoryear{Ahmad El~Sallab}{2017}]{Sallab}
Ahmad El~Sallab, Mohammed~Abdou, E.~P.
\newblock 2017.
\newblock Deep reinforcement learning framework for autonomous driving.
\newblock {\em arXiv:1704.02532}.

\bibitem[\protect\citeauthoryear{David~Silver}{2017}]{Silver}
David~Silver, Julian~Schrittwieser, K.~S.
\newblock 2017.
\newblock Mastering the game of go without human knowledge.
\newblock {\em Nature}.

\bibitem[\protect\citeauthoryear{Hado~van Hasselt}{2016}]{Hass}
Hado~van Hasselt, Arthur~Guez, D.~S.
\newblock 2016.
\newblock Deep reinforcement learning with double q-learning.
\newblock {\em Proceedings of the Thirtieth AAAI Conference on Artificial
  Intelligence (AAAI-16)}.

\bibitem[\protect\citeauthoryear{Kristofer De~Asis}{2018}]{DeAsisPer}
Kristofer De~Asis, R.~S.
\newblock 2018.
\newblock Per-decision multi-step temporal difference learning with control
  variates.
\newblock {\em arxiv:1807.01830}.

\bibitem[\protect\citeauthoryear{Kristopher De~Asis}{2018}]{DeAsis}
Kristopher De~Asis, 1 J. Fernando Hernandez-Garcia, . G. Z.~H.
\newblock 2018.
\newblock Multi-step reinforcement learning: A unifying algorithm.
\newblock {\em arXiv:1703.01327}.

\bibitem[\protect\citeauthoryear{Nan~Jiang}{2015}]{JiangDoub}
Nan~Jiang, L.~L.
\newblock 2015.
\newblock Doubly robust off-policy value evaluation for reinforcement learning.
\newblock {\em arxiv:1511.03722}.

\bibitem[\protect\citeauthoryear{Richard~Sutton}{2015}]{SutEmp}
Richard~Sutton, A. R.~M.
\newblock 2015.
\newblock An emphatic approach to the problem of off-policy temporal-difference
  learning.
\newblock {\em arxiv:1503.04269}.

\bibitem[\protect\citeauthoryear{Richard~Sutton}{2017}]{SutBar}
Richard~Sutton, A.~B.
\newblock 2017.
\newblock {\em Reinforcement Learning: An Introduction}.

\bibitem[\protect\citeauthoryear{Shashua}{2016}]{Shalev}
Shashua, S. S.-S. S. S.~A.
\newblock 2016.
\newblock Safe, multi-agent, reinforcement learning for autonomous driving.
\newblock {\em arXiv:1610.03295}.

\bibitem[\protect\citeauthoryear{Volodymyr~Mnih}{2013}]{Mnih}
Volodymyr~Mnih, Koray~Kavukcuoglu, D.~S.
\newblock 2013.
\newblock Playing atari with deep reinforcement learning.

\end{thebibliography}
\bibliographystyle{aaai}

\end{document}